\documentclass[10pt,twocolumn,letterpaper]{llncs}

\usepackage{times}
\usepackage{epsfig}
\usepackage{graphicx}
\usepackage{amsmath}
\usepackage{amssymb}

\def\abstract
   {%
   \centerline{\large\bf Abstract}%
   \vspace*{12pt}%
   \it%
   }

\usepackage{bm}
\usepackage{multirow}
\usepackage{tabularx}

\usepackage{color}
\definecolor{deepgreen}{rgb}{0,0.5,0.2}

\usepackage[final]{changes}
\definechangesauthor[color=red]{AP}
\definechangesauthor[color=blue]{APC}
\definechangesauthor[color=green]{AE}
\definechangesauthor[color=cyan]{SS}
\definechangesauthor[color=orange]{TM}

\newcolumntype{L}{>{\raggedright\arraybackslash}X}
\newcolumntype{C}{>{\centering\arraybackslash}X}
\newcolumntype{R}{>{\raggedleft\arraybackslash}X}
\newcolumntype{m}{>{$}r<{$}}	
\newcolumntype{M}{>{$}R<{$}}	
\newcolumntype{n}{>{$}c<{$}}	
\newcolumntype{N}{>{$}C<{$}}	

\newcommand{\figref}[1]{Fig{.}~\ref{#1}}	
\newcommand{\tbref}[1]{Table~\ref{#1}}
\newcommand{\secref}[1]{Sec{.}~\ref{#1}}	
\renewcommand{\eqref}[1]{Eq{.}~(\ref{#1})}	
\newcommand{\etal}{et~al{.}\ }	
\newcommand{\eg}{e{.}g{.}\ }	
\newcommand{\ie}{i{.}e{.}\ }	

\newcommand{\NetFullName}{REal-to-Synthetic Transform}
\newcommand{\NetName}{REST}  

\newcommand{\bvec}[1]{\bm{#1}}	
\newcommand{\bmat}[1]{\bm{#1}}  
\newcommand{\rfeature}{\bvec{x}^\mathrm{r}}		
\newcommand{\sfeature}{\bvec{x}^\mathrm{s}}		
\newcommand{\tsfeature}{\bvec{x}^\mathrm{ts}}	
\newcommand{\rcluster}[1]{C^\mathrm{r}_{#1}}	
\newcommand{\scluster}[1]{C^\mathrm{s}_{#1}}	
\newcommand{\sccoord}{\bvec{v}}					
\newcommand{\sccoords}{\mathcal{V}}				
\newcommand{\imcoord}{\bvec{u}}					

\newcommand{\norm}[1]{\left\| #1 \right\|}
\newcommand{\argmin}{\mathop{\rm argmin}\limits}

\newcommand{\todo}[1]{\textbf{\textcolor{green}{#1}}}

\begin{document}


\title{\NetName: Real-to-Synthetic Transform for Illumination Invariant\\Camera Localization} 

\author{
Sota Shoman\inst{1}
\and
Tomohiro Mashita\inst{1}
\and
Alexander Plopski\inst{2}
\and
Photchara Ratsamee\inst{1}
\and
Yuki Uranishi\inst{1}
\and
Haruo Takemura\inst{1}
}
\institute{
{Osaka University} \and {Nara Institute of Science and Technology}\\
\email{shoman.sota@lab.ime.cmc.osaka-u.ac.jp}
}

\maketitle

\begin{figure*}[t]
	\centering
    \includegraphics[width=\linewidth]{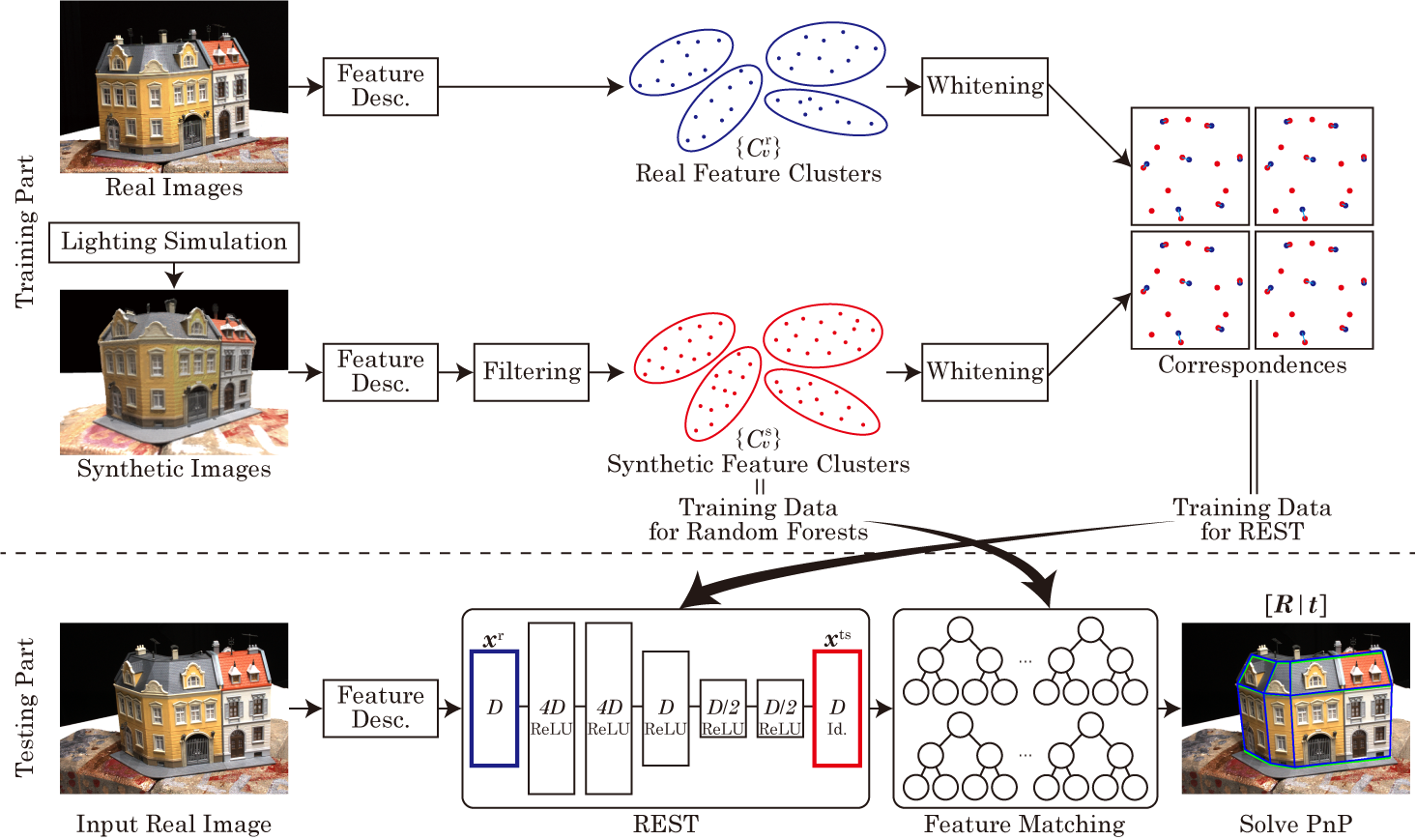}
\caption{Pipeline of our camera localization system. In the training part, lighting simulation generates synthetic images. The correspondences between features extracted from synthetic images and 3D points are obtained by ray trancing. As shown in the center, each 3D point forms synthetic feature cluster and they are used as training data for random forests. Then, real features are also described to obtain the training data for \NetName. We group real and synthetic features corresponding to the same 3D points and create feature clusters. To overlap real and synthetic feature clusters which correspond the same 3D point, whitening is applied to the both. The correspondences between a real and synthetic features are used to train \NetName. In the testing part, \NetName\ transforms input real features to synthetic features. Then, the transformed features are matched with accumulated synthetic features in the database. Finally, PnP problem is solved to obtain the camera pose.}
    \label{fig:pipeline}
\end{figure*}

\begin{abstract}

\added[id=APC]{Accurate camera localization is an essential part of tracking systems.}
\replaced[id=AP]{However, localization results are greatly affected by illumination.}{Camera localization methods are known to decrease its accuracy when illumination changes.}
\replaced[id=AP]{Including data collected under various lighting conditions can improve the robustness of the localization algorithm to lighting variation.}{To be robust to lighting variations, one of the solution is to utilize previously accumulated scenes which includes various lighting conditions.}
\replaced[id=AP]{However, this is very tedious and time consuming.}{However, accumulating various real images are time consuming.} \replaced[id=AP]{By using synthesized images it is possible to easily accumulate a large variety of views under varying illumination and weather conditions.}{Utilizing synthesized images instead of real images enables accumulating rich and controlled variations of illuminations.}
\replaced[id=AP]{Despite continuously improving processing power and rendering algorithms, synthesized images do not perfectly match real images of the same scene,}
{However current computer graphics cannot synthesize a scene completely the same as a real scene} \ie there exists a gap between real \deleted[id=AP]{images }and synthesized images \replaced[id=AP]{that}{, which} also \replaced[id=AP]{affects}{decrease} the accuracy of camera localization.
To reduce the \replaced[id=AP]{impact}{influence} of \replaced[id=AP]{this}{the} gap, we \replaced[id=AP]{introduce}{propose real-to-synthetic feature transform with an autoencoder-like network} ``\NetFullName\ (\NetName).''
\NetName\ \replaced[id=AP]{is an autoencoder-like network that converts real features to their synthetic counterpart. The converted features can then be matched against the accumulated database for robust camera localization.}{enables to match transformed features with features extracted from synthetic images.}
\replaced[id=AP]{In our experiments \NetName\ improved }{Our experiment shows that \NetName\ improve} feature matching accuracy under variable lighting conditions by approximately 30\%.
Moreover, our system outperforms state of the art CNN-based camera localization methods trained with synthetic images. \added[id=APC]{We believe our method could be used to initialize local tracking and to simplify data accumulation for lighting robust localization.}


\end{abstract}

\section{Introduction}\label{sec:intro}
Camera localization is an important task in computer vision. \replaced[id=AP]{Over the years a variety of approaches that use features extracted from the scene~\cite{Kurz_VISAPP2014,Simon_ISMAR2011}, or neural networks~\cite{Kendall_ICCV2015,Su_ICCV2015} have been developed to address it. Both approaches require a large number of images that cover the expected scene appearance to recover a representation of the scene. However, as the appearance varies during the seasons, different weather, as well as different time of the day, it is difficult and time-consuming to acquire a database that covers a sufficient variety of appearances. A cost-efficient solution is to generate synthetic views of the scene instead.}
{There exist many camera localization methods.
Most of them accumulate data which corresponds to scene 3D points to a database in advance, then match input features with features in the database to get correspondences between image points and scene 3D points, and finally solve perspective $n$-points (PnP) problem \cite{Fischler_Comm1981} to obtain the target camera pose.}
\replaced[id=AP]{Although synthetic image databases offer a lot of flexibility in deciding scene parameters, such as time of the day, weather conditions, or occluders, the synthetic scene will rarely match the appearance of the scene when captured by a camera. This is in part because it is difficult to obtain accurate geometric and optical properties for all objects in the scene. This difference inevitably leads to feature matching failure and decreased localization accuracy~\cite{Mashita_ICAT2016}.}{
Such camera localization methods has a challenge, that is sensitive to lighting conditions.
If lighting conditions are significantly changed, appearances of scene objects are also changed.
It causes failure of feature matching and decrease localization accuracy \cite{Mashita_ICAT2016}.
One of solutions is to store data under various lighting conditions to a database.
However, collecting immense camera image under various lighting conditions is impractical because it takes several months or a year to deal with time condition (morning, noon, evening or night), weather condition (sunny, cloudy, rainy or snowy) and seasonal condition (spring, summer, fall, winter).}

\deleted[id=AP]{Instead of collecting real images, we simulate various lighting scenes to generate immense synthetic images.
Synthetic images are obtained easily, but they cannot completely be the same as real images.
To generate completely the same synthetic images as real images, not only geometric shape of the scene but also optical properties of the scene are required.
However, current computer graphic techniques cannot obtain them for whole scene model.
For example, it is difficult or costs to measure reconstruct whole scene model completely the same as the actual one.
A camera can obtain the reflected light from the object but it cannot measure object color itself.
Moreover, it is possible to measure optical properties for a small model, but not for a big model such as buildings.}

\replaced[id=AP]{This problem also applies to neural nets trained on synthetic image databases. We show in this paper that a state-of-the-art CNN-based localization algorithm (PoseNet \cite{Kendall_ICCV2015}) trained on synthetic images successfully localizes synthetic views of the scene, but fails to localize actual images of the same scene.}
{As we show in this paper, PoseNet \cite{Kendall_ICCV2015} trained with synthetic images succeed camera localization from a synthetic image, but it fails camera localization from a real image.
Therefore, there exists a ``gap'' between real images and synthetic images.
The gap can cause reduction of the accuracy of feature matching between a feature extracted from a real image (called ``real feature'') and a feature extracted from a synthetic image (called ``synthetic feature'').}

\replaced[id=AP]{In this paper, we introduce an autoencoder-like network ``\NetFullName\ (\NetName)'' to overcome the ``gap'' between real images and synthetic images for feature-based localization. 
\deleted[id=SS]{We focus on feature-based localization as it is less memory intensive and thus still the most likely method used on mobile devices. \todo{RF consumes a large memory. (SS)}}}
{To be robust to the gap, we propose an autoencoder-like neural network ``\NetFullName\ (\NetName).''}
\NetName\ transforms a \deleted[id=AP]{real }feature \added[id=AP]{descriptor }extracted from an input image \added[id=AP]{(we refer to it as ``real feature'') }into a \replaced[id=AP]{corresponding feature descriptor, if it were extracted from synthetic image (we refer to it as ``synthetic feature'') that was generated}{synthetic feature which is predicted to be} under the same conditions, \ie 3D points and lighting conditions, as the real \replaced[id=AP]{image}{feature}. \replaced[id=AP]{After \NetName\ transforms a real feature it closely resembles the corresponding synthetic feature, thus increasing the accuracy of matching it with the feature database.}{
Since transformed synthetic features can be regarded as synthetic features, feature matching between transformed synthetic features and synthetic features in a database is expected to be succeed.}

\replaced[id=AP]{\NetName\ can be trained with a small number of real images, and corresponding synthetic views. That is because we do not learn how to accurately estimate the pose of the camera under varying lighting conditions, but the relationship between an extracted real feature and its corresponding synthetic feature. We evaluate our method using images from the DTU robot image dataset~\cite{Jensen_CVPR2014} that stores image of the same model taken under different lighting conditions within a 1 m $\times$ 1 m area. In this paper, we use only 48 images to train the network and achieve an accuracy of 1.66 cm and 1.41$^{\circ}$.}
{To train \NetName, correspondences between a real feature and a synthetic feature as training data, \ie real images are required for training \NetName, but the number of real images is less than that for training other CNNs.
The reason is that \NetName\ learn only the relationship between a real feature and a synthetic feature, in other words it does not need to learn illumination invariant camera localization.
As we stated above, immense real images are required to be robust lighting variations and collecting immense real images are impractical.
However, since \NetName\ does not learn illumination invariant camera localization, the number of real images can decrease.
Thus, collecting real images for training \NetName\ is practical.}

Our main contributions are the following.
\begin{itemize}
\item\added[id=AP]{We introduce \NetName\ to improve matching of real features and a database of synthetic features.}
\item
\replaced[id=AP]{We show that even when trained on a small number of images \NetName\ improves localization accuracy compared to state-of-the-art methods trained only on synthetic images.}{We show that \NetName\ improves feature matching accuracy with machine learning trained with synthetic features.}
\item 
\replaced[id=AP]{We show that \NetName\ is robust to illumination changes and its performance with different feature descriptors.}{
We realize illumination invariant camera localization using less real images than other CNN-based camera localization methods.}
\end{itemize}

\section{Related Work}\label{sec:related_work}
\replaced[id=APC]{Over work focuses on camera localization from synthetic images. In this section we review existing methods on localization and discuss how they utilized synthetic images.}
{Our goal is illumination invariant camera localization and to reduce the influence of the gap between real images and synthetic images. In this section, we introduce state of the art camera localization methods and simulation methods.} \replaced[id=AP]{Camera localization methods~\cite{Kurz_VISAPP2014,Simon_ISMAR2011,Kendall_ICCV2015,Mashita_ICAT2016,Kurz_ISMAR2012,Shinozuka_IMVIP2014,Newcombe_ISMAR2011,Kerl_IROS2013,Engel_ICCV2013,Engel_ECCV2014,Kendall_ICRA2016,Walch_ICCV2017} can be divided into}
{There exist many camera localization approaches \cite{Mashita_ICAT2016,Kendall_ICCV2015,Kurz_ISMAR2012,Kurz_VISAPP2014,Shinozuka_IMVIP2014,Simon_ISMAR2011,Newcombe_ISMAR2011,Kerl_IROS2013,Engel_ICCV2013,Engel_ECCV2014,Kendall_ICRA2016,Walch_ICCV2017}.
They can be divided to three types;} feature-based method\added[id=AP]{s}\deleted[id=AP]{, simultaneous localization and mapping (SLAM)} and convolutional neural network (CNN)-based method\added[id=AP]{s}. \added[id=AE] {We discuss each category as follow.}

\paragraph{Feature-based Camera Localization.}
\replaced[id=AP]{Feature-based methods~\cite{Kurz_VISAPP2014,Simon_ISMAR2011,Mashita_ICAT2016,Kurz_ISMAR2012,Shinozuka_IMVIP2014} detect 3D points stored in a database in the camera image and estimate the camera pose by solving the perspective $n$-point problem~\cite{Fischler_Comm1981}. To detect a 3D point in the image these methods extract a feature descriptor for the detected feature (corner, blob, etc.) in the camera image and for each 3D point match it with the corresponding descriptor stored in the database. When a good match is detected the 2D feature is matched to the corresponding 3D point. Over the years, a variety of descriptors, \eg SIFT \cite{Lowe_IJCV2004}, SURF \cite{Bay_ECCV2006}, ORB \cite{Rublee_ICCV2011} and LIFT \cite{Yi_ECCV2016}, that are robust to constrained viewpoint changes have been developed. Storing descriptors for all possible 3D points is not viable, as many features can be discovered only under certain conditions, it leads to a very large database and consequently very long processing time.} {
Feature-based approaches \cite{Kurz_ISMAR2012,Kurz_VISAPP2014,Mashita_ICAT2016,Shinozuka_IMVIP2014,Simon_ISMAR2011} use feature matching between image pixels and some reference points in the scene. Many feature descriptors are proposed, \eg SIFT \cite{Lowe_IJCV2004}, ORB \cite{Rublee_ICCV2011} and LIFT \cite{Yi_ECCV2016}. }

\added[id=AP]{Feature databases can be obtained through Structure-from-Motion reconstruction~\cite{Ma_Springer}, Simultaneous Localization and Mapping methods~\cite{Newcombe_ISMAR2011,Kerl_IROS2013,Engel_ICCV2013,Engel_ECCV2014,MurArtal_TRo2015,Tateno_CVPR2017}, geo-annotated image datasets~\cite{Rattenbury_SIGIR2007}, or 3D scans of the environment~\cite{Kurz_VISAPP2014,Kurz_ISMAR2013}. As all of these methods reconstruct the scene appearance at a particular point in time, they are not robust to drastic changes in lighting conditions. To account for these changes, the database must be generated multiple times under different conditions.}

\deleted[id=AP]{Most of feature descriptors are robust to scale and rotation, but if viewpoints are significantly changed, feature matching accuracy is decreased.}
\replaced[id=AP]{A large database often contains a number of redundant, rarely seen features, many similar features, and multiple descriptors for the same feature. To reduce the size of the database a ``representative subset'' can be used to ensure good localization results~\cite{Kurz_VISAPP2014,Kurz_ISMAR2012}. A subset contains \replaced[id=SS]{a particular number of the}{$N$} most representative descriptors. These descriptors are either selected based on feature visibility in different views, or descriptor matching robustness. By removing redundant feature descriptors, and reducing the number of overall features in the database these subsets reduce the time required for localization, and improve the robustness of the matching.}{
Kurz \etal \cite{Kurz_ISMAR2012,Kurz_VISAPP2014} select some ``representative features'' which is invariant to viewpoint.
This enables camera localization to be faster.}
\replaced[id=AP]{Feature matching robustness can also be improved by taking advantage of the various sensors embedded into mobile devices~\cite{Kurz_VISAPP2014,Simon_ISMAR2011}.}
{To be robust to viewpoint changes, \cite{Kurz_VISAPP2014,Simon_ISMAR2011} use some sensor data, \eg GPS, accelerometers, compass and inertial sensor.}
\replaced[id=SS]{Different methods \cite{Simon_ISMAR2011,Mashita_ICAT2016,Shinozuka_IMVIP2014} also utilize synthetic images for viewpoint invariant camera localization.}
{\cite{Simon_ISMAR2011,Shinozuka_IMVIP2014,Mashita_ICAT2016} propose viewpoint invariant camera localization methods using synthetic images.}

\deleted[id=AP]{Many types of SLAM methods are proposed.
Since we aimed to monocular camera localization, we review some monocular SLAM methods.
Monocular SLAM methods can be divided into three types;
direct monocular SLAM \cite{Newcombe_ISMAR2011,Kerl_IROS2013,Engel_ICCV2013,Engel_ECCV2014}, feature-based monocular SLAM \cite{MurArtal_TRo2015}, and CNN-based monocular SLAM \cite{Tateno_CVPR2017}.}

\deleted[id=AP]{Large Scale Direct SLAM (LSD-SLAM) \cite{Engel_ECCV2014} uses semi-dense maps which is computed by gradient of the input image.
This enables to reconstruct large scale environment in real-time on a CPU\@.
ORB-SLAM \cite{MurArtal_TRo2015} is one of state of the art feature-based SLAM in real-time on CPU\@.
This method is based on ORB features \cite{Rublee_ICCV2011}, which enables to optimize key frames by bundle adjustment.
CNN-SLAM \cite{Tateno_CVPR2017} efficiently fuses CNN-predicted dense maps with depth estimation computed by direct monocular SLAM\@.
This enables to obtain depth in texture-less region, and to estimate the absolute scale of the reconstruction.}

\deleted[id=AP]{SLAM is able to be create a scene map which enables to many tasks in computer vision, \eg augmented reality and robot navigation.
However, the map is not robust to a significant change of lighting conditions.
Thus, it is difficult to reuse the generated maps under lighting variable environment.}

\paragraph{CNN-based Camera Localization.}
CNNs have been deployed for many tasks in computer vision, especially image classification~\cite{Krizhevsky_NIPS2012,Szegedy_CVPR2015}.
\replaced[id=AP]{With the increasing number of available images in various databases, localization with CNNs has gained a lot of attention recently~\cite{Kendall_ICCV2015,Kendall_ICRA2016,Walch_ICCV2017}. As these methods require many images with known camera pose, some systems use synthetic data to estimate viewpoints~\cite{Su_ICCV2015} and to predict 6D object poses~\cite{Kull_ICCV2015}. To overcome the gap between synthetic and real images, synthetic images can be post-processed with a generative adversarial network (GAN) to change their appearance to more closely resemble that of real images~\cite{Shrivastava_CVPR2017}. However, training a GAN requires between 10K and 100K image pairs. It is thus difficult to obtain sufficient number images of the scene under varying lighting conditions to train a GAN.}
{Recently, CNN-based Camera Localization methods~\cite{Kendall_ICCV2015,Kendall_ICRA2016,Walch_ICCV2017} are proposed.
PoseNet \cite{Kendall_ICCV2015} is state of the art CNN-based camera localization method, which applies GoogLeNet \cite{Szegedy_CVPR2015} for camera localization.
There are some extensions of PoseNet including Bayesian PoseNet \cite{Kendall_ICRA2016} and a method of Walch \etal \cite{Walch_ICCV2017}.}

\deleted[id=AP]{Using synthesized data is one of solutions for training deep neural networks (DNNs).
As it is well known, DNN runs fast and performs well but requires large amount of data for training.
Namely, DNN cannot demonstrate good performance if enough number of data are unavailable.
Synthetic data is one of solutions for this dataset problem in DNNs.}

\deleted[id=AP]{Using CNNs trained with synthetic images, \cite{Su_ICCV2015} estimates viewpoints and \cite{Kull_ICCV2015} predicts object 6D poses.
\cite{Su_ICCV2015} uses synthetic images rendered from different camera pose under a various lighting conditions. To avoid overfitting, these rendered images are composed smoothly with background and reduce high contrast object boundaries.
\cite{Kull_ICCV2015} solves object pose estimation problem by minimizing energy which is computed by a CNN. It inputs rendered data from a 3D model and observed data from an input RGB-D image to the CNN.
SimGAN \cite{Shrivastava_CVPR2017} is also a famous method using synthetic images, which generates a refined image from a synthetic image using a generative adversarial network (GAN) trained with unlabeled real images and labeled synthetic images.
Since the generated images can be regarded as real images, we can use them to train another machine learning architecture.
However, to train a GAN, from 10K to 100K real images are required.
As we stated in \secref{sec:intro}, collecting many real images under a various lighting conditions are  impractical.}

\deleted[id=TM]{Illumination robust localization requires large image databases recorded at different times to ensure sufficient variability. While it is possible to generate these databases from synthetic images, these images differ from real images due to geometrical and optical differences of the real and virtual objects. While it is possible to adjust synthetic images so they similar to real images through the use of GAN, these also require large databases to ensure good conversion. We overcome this problem by converting only the relevant part of the image (feature descriptors). While the idea behind our method is similar to \cite{Shrivastava_CVPR2017}, we use a more simple approach and thus require far less image pairs to train our network.}
\deleted[id=AP]
{In this section, we explained that feature descriptors are sensitive to lighting variations,
which causes that scene maps generated with SLAM cannot be reused.
In addition, using synthetic images can solve dataset problem in DNNs.
SimGAN \cite{Shrivastava_CVPR2017} can solve the gap problem, but it is not suitable to variable lighting problem because collecting real data under various lighting conditions is difficult.}

\deleted[id=AP]{In this paper, we will illustrate that these problem above can be solved.
For variable lighting problem, illumination simulation can generate immense synthesized data.
Since this improves feature matching accuracy, scene maps generated with SLAM can be reused.
When scene maps are available, feature matching between two synthesized data, \ie variable lighting problem can be solved in synthesized data.
However, due to the gap problem, feature matching between real data and synthesized data is difficult.
Then, transformation from real data to synthesized data enables feature matching.
Since SimGAN requires large number of real data for training, we utilize autoencoder to transform a real feature to a synthetic feature.
Our method requires less real data than SimGAN.}

\deleted[id=TM]{Real-to-Synthetic transform is more beneficial than synthetic-to-real transform. As we stated in \secref{sec:intro}, camera localization is solved by feature matching between input features and database features. Lighting simulation enables to solve this feature matching problem as a pattern recognition problem because we know the parameters of the simulation environment. Thus, the higher density of features, the more accuracy of feature matching is improved. Hereby, there exists two pattern transforms, real-to-synthetic transform and synthetic-to-real transform. Factors of real scene, \eg light sources, reflections etc{.}, are much more than that of simulation scene. Some factors of real scene are difficult to be measured. Real-to-synthetic transform can limit the number of factors, which enables to reduce the impact of such uncontrollable factors. On the other hand, synthetic-to-real transform cannot limit the number of factors. To match features correctly, interpolation for such uncontrollable factors are needed, which requires more real data. Thus, real-to-synthetic transform is more beneficial in camera localization.}


\section{Real-to-Synthetic Feature Transform}
Figure~\replaced[id=SS]{\ref{fig:pipeline}}{\ref{fig:problem_statement}} shows \added[id=AP]{an }overview of our camera localization system.
Our \deleted[id=AP]{final }goal is to estimate the camera pose $[\bmat{R} | 
\bvec{t}]$, where $\bmat{R}$ is a rotation matrix and $\bvec{t}$ is a translation vector, from a single input RGB image.

\replaced[id=AP]{We generate synthetic images that simulate a large variety of lighting conditions.}
{To be robust to lighting conditions, we first execute a illumination simulation and generate synthetic images.}
\replaced[id=AP]{Afterward, features will be extracted from these synthetic views and compute the corresponding 3D location from known camera parameters and scene model. For each features we store the computed 3D location and descriptor in a database.}
{Then, synthetic features are extracted from the synthetic images and recorded to a database with corresponding scene coordinates. Since synthetic images are rendered by \deleted[id=AP]{virtual} virtual camera whose parameters are known, scene coordinates for each pixels can be computed.}

\replaced[id=AP]{To estimate the pose of an input image, we detect real features $\rfeature$ and match them with our database of synthetic features $\sfeature$. From the matched 2D-3D points, we compute the camera pose with the \replaced[id=SS]{perspective $n$-point (PnP)}{PnP} algorithm~\cite{Fischler_Comm1981}. If the number of correct matches is too small, due to the gap between descriptors acquired from real and synthetic images, the pose estimation fails to recover the correct pose.}
{Hereby, we have real features $\rfeature$ extracted from an input image and synthetic features $\sfeature$ extracted from synthetic images generated by illumination simulation.
Camera pose can be computed by PnP algorithm \cite{Fischler_Comm1981} from feature matching results.
If we match a real feature with a synthetic feature, almost all matching is failed due to the gap between them.
We call this problem ``gap problem,'' and we aim to solve this problem.}

\added[id=AP]{Given identical light conditions, the gap between real and synthetic features is due to incorrect assumptions about the geometry and optical properties of the scene.}
\added[id=SS]{We explain how to overcome these inaccuracy in this section.}

\subsection{Synthetic-to-Real vs. Real-to-Synthetic}\label{sec:vs}
\added[id=TM]{As discussed in the Sec. 1, utilizing image synthesis is a solution for the scene variation problem however we need feature transform between synthetic and real feature. There are two possible ways of feature transform which are real-to-synthetic and synthetic-to-real respectively. An obvious difference between real and synthetic scene is in their complexity. Generally, a real scene is very complex and vary with a number of factors. Moreover, some of the factors are difficult to measure, estimate, and/or synthesize. In other words, a synthetic scene is a simplification of the original real scene (i.e. a subset of real scene). Therefore, synthetic-to-real transform needs latent variables to supplement the difference in complexity. SimGAN \cite{Shrivastava_CVPR2017} does not use latent variables which shows a potential of synthetic-to-real transform. However, this method requires a large amount of real images and learning of a large network for representing complex real scene. 

\indent On the other hand, real-to-synthetic transform can be achieved with a simpler network because it is a simplification of the real scene and does not require latent variables.
We therefore adopt real-to-synthetic transform as our camera localization. Moreover, our real-to-synthetic transform, \NetName, is learned with small number of real images.  
}


\added[id=AP]{\replaced[id=SS]{We hereby execute real-to-synthetic feature transform with an autoencoder-like network \NetName\ to overcome the gap between real and synthetic information.}
{We overcome these inaccuracies with an autoencoder-like network \NetName\added[id=SS]{\ that transforms a real feature into a synthetic feature}.} 
Autoencoders have been used for dimension compression~\cite{Hinton_Science2006}, pre-training of deep neural networks and denoising~\cite{Vincent_JMLR2010}. We consider the gap between real and synthetic features to be due to noise and train \NetName\ to minimize the loss function}
\begin{equation}
L(\tsfeature) = \norm{ \tsfeature - {\sfeature} },\label{eq:loss}
\end{equation}
\added[id=AP]{\noindent where $\tsfeature$ is the real feature $\rfeature$ transformed with \NetName, and ${\sfeature}$ is the corresponding ground truth synthetic feature.}
\added[id=SS]{To train \NetName\ efficiently, we pre-train \NetName\ to transform a synthetic feature into itself before the main training with correspondences between real and synthetic features.}

\added[id=TM]{
One of the disadvantages of \NetName\ is in expansion of map and database. To apply \NetName\ to expanded a map and a database, \NetName\ must be learned from scratch. In the case of synthetic-to-real transform, theoretically its expandability is higher because new feature can be directly accumulated to the database. 
}

\deleted[id=AP]{To train \NetName, correspondences between a real feature and a synthetic feature are required.
We show how to predict the correspondences in \secref{sec:training_data}.}

\deleted[id=AP]{Autoencoder is a kind of neural networks, whose the number of input units and that of output units are the same.
Examples of applications of autoencoder are compression of dimension \cite{Hinton_Science2006}, pre-training of deep neural networks and denoising \cite{Vincent_JMLR2010}.
Autoencoder for denoising is trained to convert a noise-added image into the original image.
We apply autoencoder to transform a real feature to a synthetic feature.}

\deleted[id=AP]{Given the correspondences between a real feature and a synthetic feature, \NetName\ is trained by minimizing a loss function,}
\deleted[id=AP]{where $\tsfeature$ is transformed synthetic feature from a real feature $\rfeature$ with \NetName, and ${\sfeature}$ is ground truth synthetic feature corresponding to the real feature $\rfeature$.}

\subsection{Correspondences between Real and Synthetic Features}\label{sec:correspondence}
\replaced[id=SS]{To train \NetName, correspondence between real and synthetic features are required. Ideally the correspondences between real and synthetic feature both of which are under the same lighting condition are the best, but this is difficult as discussed above. However, by using projection matrices, the correspondence between a set of real features and a set of synthetic features both of which correspond the same 3D point. We use the set correspondences to obtain the feature correspondences. Hereby, we assume that some real images $\{ I_n \}_{n=1}^\mathrm{N}$ and their corresponding projection matrices $\{ \bmat{P}_n \}_{n=1}^\mathrm{N}$ are known, where $\mathrm{N}$ represents the number of real images for training.}
{To train \NetName, a dataset which has correspondences between each real features and synthetic features. Hereby, we consider given some real images $\{ I_n \}_{n=1}^\mathrm{N}$ and corresponding projection matrices $\{ \bmat{P}_n \}_{n=1}^\mathrm{N}$, where $\mathrm{N}$ represents the number of real images for training.}
The making of correspondences is separated to three steps below.

First, we compute 3D points for each real and synthetic features.
3D points of synthetic features can be computed by ray casting because we know the camera parameters used in a simulation.
Let $\sccoords$ denotes the set of the 3D points of synthetic features.
Then, a 3D point of a real feature $\rfeature$\added[id=SS]{,} which is extracted from a real image $\mathrm{I}_n$\added[id=SS]{,} are searched from $\sccoords$ by
\begin{equation}
	\sccoord = \argmin_{\sccoord' \in \sccoords} \norm{ \pi \left( \bmat{P}_n \dot{\sccoord}' \right) - \imcoord },
\end{equation}
where $\pi$ represents perspective projection function defined by
\begin{equation}
	\pi\left( \left[ x, y, \lambda \right] \right) = \left[ \frac{x}{\lambda}, \frac{y}{\lambda} \right],
\end{equation}
$\dot{\sccoord}'$ represents a homogeneous vector of $\sccoord'$ and $\imcoord$ represents image points of the real feature $\rfeature$.

Secondly, we group each features by 3D points.
We call the group of real features corresponding to a 3D point $\sccoord$ ``real feature cluster'' denoted by $\rcluster{\sccoord}$ and call the group of synthetic features corresponding to a 3D point $\sccoord$ ``synthetic feature cluster'' denoted by $\scluster{\sccoord}$.
We match $\rcluster{\sccoord}$ with $\scluster{\sccoord}$.

\begin{figure}[t]
	\centering
    \includegraphics[width=\linewidth]{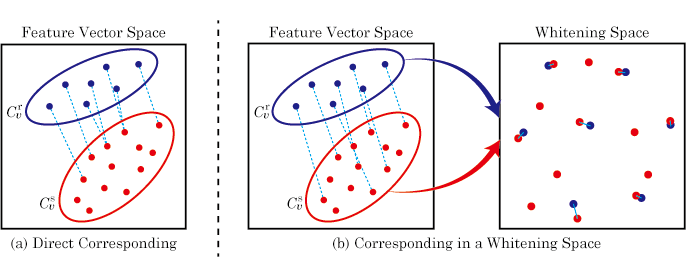}
    \caption{\replaced[id=SS]{Corresponding between $\rfeature$ in $\rcluster{\sccoord}$ and $\sfeature$ in $\scluster{\sccoord}$.}{Matching a real feature in a real feature cluster $\rcluster{\sccoord}$ with a synthetic feature in a synthetic feature cluster $\scluster{\sccoord}$ corresponding to the same \replaced[id=SS]{3D point}{scene coordinate} $\sccoord$.} \deleted[id=SS]{Red ellipses and blue ellipses represents real feature cluster and synthetic feature cluster. }Red \deleted[id=SS]{dots }and blue dots represents real features and synthetic features\replaced[id=SS]{ respectively. Each ellipses represents a cluster.}{.}}
    \label{fig:feature_matching}
\end{figure}

Finally, we \replaced[id=SS]{make correspondences between a real feature $\rfeature \in \rcluster{\sccoord}$ and a synthetic feature $\sfeature \in \scluster{\sccoord}$.}{match a real feature $\rfeature \in \rcluster{\sccoord}$ with a synthetic feature $\sfeature \in \scluster{\sccoord}$.}
Since the distribution of a real feature cluster in the feature vector space and that of a synthetic feature cluster in the feature vector space are expected to be different, \eg shifted or scaled.
\replaced[id=SS]{If we directly correspond a $\rfeature$ with a $\sfeature$ in $\scluster{\sccoord}$ under a positional relationship like \figref{fig:feature_matching}(a), most of $\rfeature$ will be corresponded with $\sfeature$ located in boundary side of the $\scluster{\sccoord}$. As a result of the learning with this correspondences, $\rfeature$ will be transformed to a position close to the boundary with another cluster and it will increase failure matching.}
{If \replaced[id=SS]{making correspondences}{matching} directly, most real features are matched with synthetic features near the boundary of the synthetic feature cluster as shown in \figref{fig:feature_matching}(a). Most of feature matching approaches use decision boundaries to classify, so features near decision boundaries may be mismatched, \ie direct matching can cause decrease of feature matching accuracy.}
To \replaced[id=SS]{make better correspondences between a real and synthetic feature}{match a real feature with a synthetic feature correctly}, we apply whitening transformation \cite{Krizhevsky_TechnicalReport2009} to both of a real \deleted[id=SS]{feature cluster }and a synthetic feature cluster as shown in \figref{fig:feature_matching}(b). Since whitening changes the mean into zero and the variance into one, in the whitening space two clusters are overlapped.
Then, we deploy nearest neighbor in the whitened space to obtain the correspondences between a real feature and a synthetic feature.
\deleted[id=SS]{Using this correspondences, most real features are not transformed near the boundary of corresponding synthetic feature cluster, thus this mapping does not cause the decrease of feature matching accuracy.}

\subsection{Preparation of Training Data for \NetName}\label{sec:train_rest}
\added[id=SS]{As stated in \secref{sec:correspondence}, we can obtain correspondences between real and synthetic features, but this is not enough to train \NetName\ because the number of real features are smaller than that of synthetic features. Small number of training data cannot train \NetName\ enough. Additionally they cause to transform a real feature into a outlier, which decreases the accuracy of feature matching. Thus the data augmentation is required.}

\deleted[id=SS]{In this section we illustrate the detail of \NetName.
As shown in \figref{fig:pipeline} \NetName\ has \replaced[id=SS]{six}{6} layers except for the input layer.
Since the mapping from a real feature a synthetic feature is expected to be non-linear, we use ReLU as an activation function of hidden layers.}

For \added[id=SS]{the }preparation of the \added[id=SS]{sufficient }training data, we leverage $k$-nearest-neighbor ($k$-NN) in order to increase the number of correspondences between a real and synthetic feature for training data because we use \replaced[id=SS]{a small number of}{not many} real images for training, \ie the number of real features extracted from them is \replaced[id=SS]{less than that of synthetic features}{not many}.
Moreover, since we train \NetName\ with $k$ correspondences $(\rfeature, \sfeature_1), \cdots, (\rfeature, \sfeature_k)$ per a real feature $\rfeature$ and we use mean square error defined by \eqref{eq:loss} as the loss function, $\rfeature$ is transformed to the mean of $\{ \sfeature_1, \cdots, \sfeature_k \}$.
This means that $\rfeature$ transformed more far from the boundaries of the corresponding synthetic cluster.
As we stated in \secref{sec:correspondence}, if a real feature is transformed\added[id=SS]{ into} near the boundaries\replaced[id=SS]{, it will be possible to cause matching failure.}{make feature matching difficult.}
\deleted[id=SS]{Thus, using $k$-NN is expected to improve matching accuracy.}

\deleted[id=SS]{Since we assume that the number of real images are much less than that of synthetic images, the number of the correspondence between a real feature and a synthetic feature is not many. Hereby, we utilize $k$-NN to get $k$ correspondences per a real feature. Thus, we can obtain enough number of training data.}

Additionally, using $k$-NN enables to train \NetName\ effectively.
We have representative features in a database.
\replaced[id=SS]{Since representative features \cite{Kurz_ISMAR2012} are robust to camera viewpoint and rotation, the more synthetic features a synthetic cluster has, the more robust to camera viewpoint and rotation such synthetic features are. Therefore focusing on training to transform a real feature into such a synthetic feature improves feature matching accuracy.}
{As stated in \secref{sec:related_work}, representative features \cite{Kurz_ISMAR2012} are robust to camera view\added[id=SS]{point and rotation}. Thus, a synthetic cluster which has many representative features are corresponding to a \replaced[id=SS]{3D}{reference} point which is robust to camera view, \ie \replaced[id=SS]{the synthetic features in such a cluster are}{is} matched easily.}
\replaced[id=SS]{To focus on such training}
{To focus on transforming\added[id=SS]{ a real feature in}to \replaced[id=SS]{such a synthetic feature}{such a synthetic features}}, we optimize the $k$ value by
\begin{equation}
	k = \left\lfloor \gamma \sqrt{|\scluster{\sccoord}|} \right\rfloor, \label{eq:knn}
\end{equation}
where $\gamma$ denotes a parameter to adjust $k$ and $|\scluster{\sccoord}|$ denotes the number of synthetic features which $\scluster{\sccoord}$ contains.

\section{\replaced[id=SS]{Implementation}{Lighting Robust Camera Localization from Synthetic Images}}\label{sec:implementation}

In this section, we illustrate our lighting robust camera localization system.
\deleted[id=AP]{Figure~\ref{fig:pipeline} shows the pipeline.}
Our system \replaced[id=AP]{consists of a}{has} training and testing part\added[id=AP]{ shown in \figref{fig:pipeline}}. \replaced[id=AP]{Training includes the following steps:}{In the training part,}
\begin{enumerate}
	\item \replaced[id=AP]{G}{To g}enerate synthetic images\deleted[id=AP]{, execute a simulation} under various lighting conditions. We \deleted[id=AP]{will }explain the detail\added[id=AP]{s} in \secref{sec:simulation}.
    
    \item Extract synthetic features from \replaced[id=AP]{these images and select a representative subset as described in~\cite{Kurz_ISMAR2012}.}{synthetic images and apply \cite{Kurz_ISMAR2012} to select features which are robust to camera view. Selected synthetic features are recorded to a database.}

	\item To prepare the training data of \NetName, we extract real features from some real images, make real and synthetic feature clusters, and \replaced[id=SS]{find correspondence between real and synthetic features}{match real features with synthetic features} as stated in \secref{sec:correspondence}. \added[id=SS]{Moreover we utilize $k$-NN to augment the training data as stated in \secref{sec:train_rest}. We empirically set the parameter $\gamma$ in \eqref{eq:knn} to 0.2.}

	\item Train \NetName\added[id=AP]{ that transforms extracted features $\rfeature$ to $\tsfeature$}\added[id=SS]{ using the loss function \eqref{eq:loss}}. \deleted[id=SS]{We \deleted[id=AP]{will }explain the detail\added[id=AP]{s} in \replaced[id=SS]{\secref{sec:train_rest}}{\secref{sec:train_net}}\added[id=AP]{.}} 

	\item Train a random forest \replaced[id=AP]{that}{which} matches a transformed feature \added[id=AP]{$\tsfeature$ }with \replaced[id=AP]{the database}{a synthetic feature}.
\end{enumerate}
In the testing part,
\begin{enumerate}
	\item Extract real features \added[id=AP]{$\rfeature$ }from the input \deleted[id=AP]{real }image.
    \item Transform \replaced[id=AP]{$\rfeature$ into $\tsfeature$}{the real features} by \NetName.
    \item Match \replaced[id=AP]{$\tsfeature$}{the transformed features} with \replaced[id=AP]{the database}{synthetic features} by the trained random forest.
    \item Solve PnP problem using random sample consensus (RANSAC) \replaced[id=AP]{to}{and} obtain the camera pose.
\end{enumerate}

\subsection{Simulation under Various Lighting Conditions}\label{sec:simulation}
\begin{figure}[t]
	\centering
	\begin{minipage}{0.45\linewidth}
		\includegraphics[width=\linewidth]{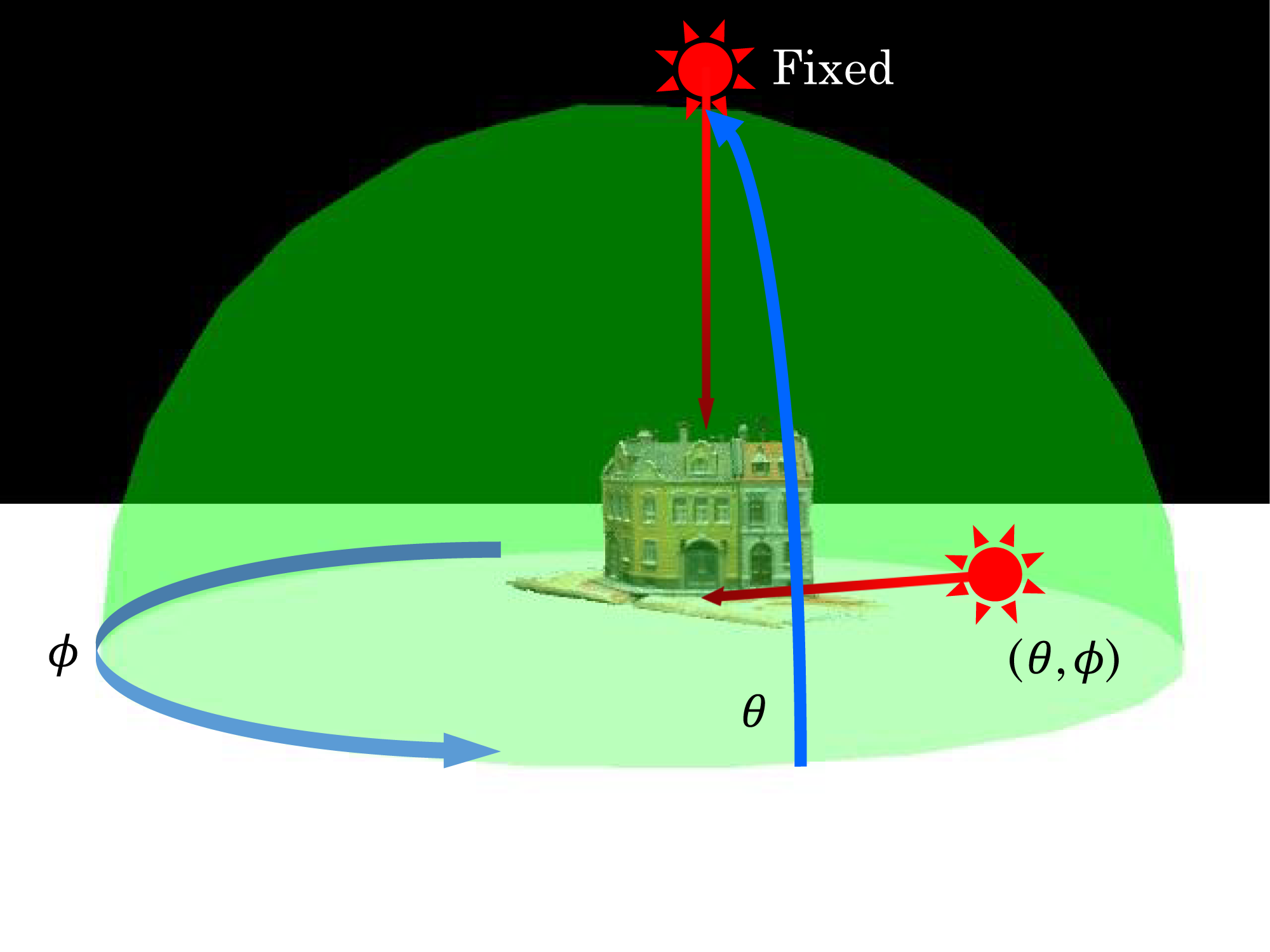}
        \caption{Light sources used in illumination simulation.}
        \label{fig:light_source}
	\end{minipage}
    \hspace{1em}
    \begin{minipage}{0.45\linewidth}
    	\includegraphics[width=\linewidth]{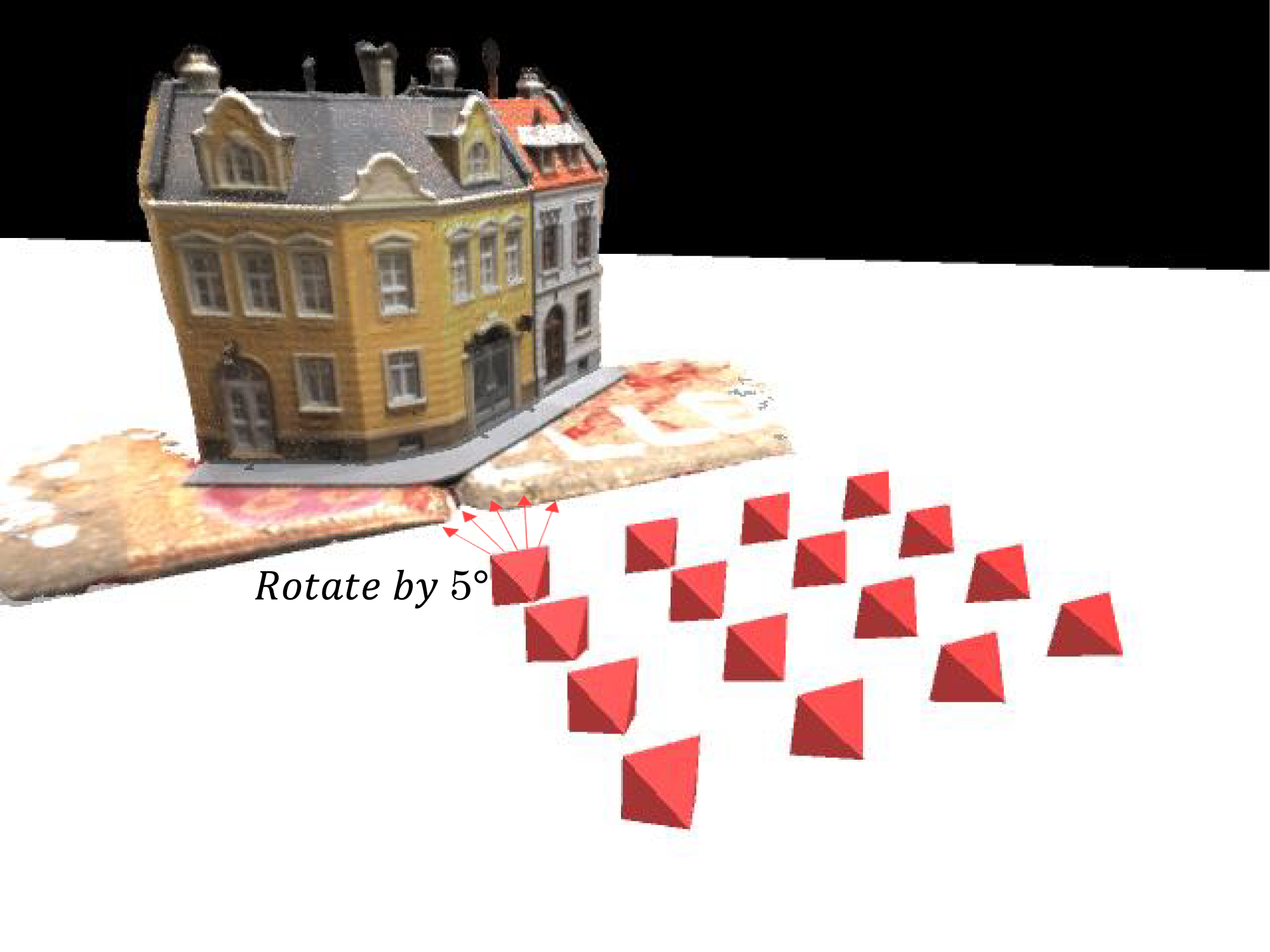}
        \caption{Camera positions for generating synthetic images.}
        \label{fig:camera_position}
    \end{minipage}
    \\
    \vspace{2ex}
	\newlength{\imgwidth}
    \setlength{\imgwidth}{.2\linewidth}
	\centering
    {\tabcolsep=0ex
    \begin{tabular}{ccccc}
    	\includegraphics[width=\imgwidth]{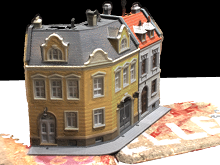}
        & \includegraphics[width=\imgwidth]{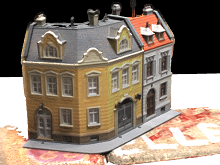}
        & \includegraphics[width=\imgwidth]{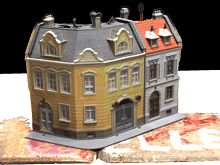}
        & \includegraphics[width=\imgwidth]{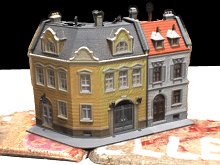}
        & \includegraphics[width=\imgwidth]{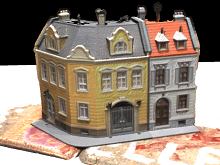}\\
        
        \includegraphics[width=\imgwidth]{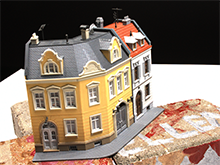}
        & \includegraphics[width=\imgwidth]{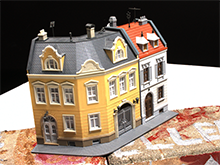}
        & \includegraphics[width=\imgwidth]{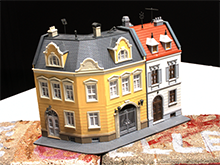}
        & \includegraphics[width=\imgwidth]{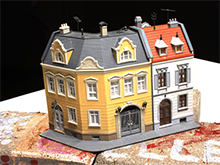}
        & \includegraphics[width=\imgwidth]{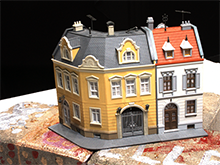}
    \end{tabular}
    }
    \caption{Rendered synthetic images (top) and real images (bottom).}
    \label{fig:synthetic_images}
\end{figure}

\replaced[id=AP]{To simulate the scene appearance under varying lighting conditions, we require a 3D model which can be obtained through several approaches such as structure from motion, dense tracking and mapping (DTAM) \cite{Newcombe_ICCV2011}, dense environment scans~\cite{Kurz_VISAPP2014,Kurz_ISMAR2013}, or Poisson surface reconstruction \cite{Kazhdan_SGP2006,Calakli_CGF2011,Kazhdan_TOG2013}.}
{To generate synthetic images under various lighting conditions, we simulate some lighting environment.
Our approach requires scene models for simulation.
There exist many 3D reconstruction methods, \eg structure from motion, dense tracking and mapping (DTAM) \cite{Newcombe_ICCV2011}, Poisson surface reconstruction \cite{Kazhdan_SGP2006,Calakli_CGF2011,Kazhdan_TOG2013}.}
\replaced[id=AP]{In this paper}{For the scene object model,} we use the DTU robot image dataset \cite{Jensen_CVPR2014}.
This data\deleted[id=AP]{ }set contains \added[id=AP]{a }structure-from-motion model, real images taken under different lighting conditions, and \added[id=AP]{corresponding }projection matrices\deleted[id=AP]{ corresponding to each images}.
\replaced[id=AP]{We obtain a mesh model through Poisson surface reconstruction \deleted[id=SS]{\cite{Kazhdan_SGP2006,Calakli_CGF2011,Kazhdan_TOG2013}} of the structure-from-motion model.}
{To obtain the mesh model, we apply Poisson surface reconstruction \cite{Kazhdan_TOG2013} to the structure-from-motion model.}

\replaced[id=AP]{To match the illumination conditions in the DTU dataset, we use two parallel light sources. As presented in \figref{fig:light_source}, one light source illuminates the scene from above, while the other is moved around the scene and always oriented towards the model, as in~\cite{Jensen_CVPR2014}. }
{For illumination simulation, we use two parallel light source in order to be the same condition as \cite{Jensen_CVPR2014}. One illuminates from the top of the model downward. The other illuminates from surrounding of the model to the center of the model as shown in \figref{fig:light_source}.
We change the direction of the dynamic light source.}
\replaced[id=AP]{The direction the light is coming from }
{The direction }is represented by latitude angle $\theta$ and longitude angle $\phi$.
$\theta$ is set to $\{ 10, 20, 30, \cdots, 80 \}$ and $\phi$ is set to $\{0, 30, 60, \cdots, 180 \}$ degrees, \ie we simulate \deleted[id=AP]{under }56 lighting conditions in total.

\replaced[id=AP]{We generate views of the scene from $4\times4$ positions in front of the model (shown in Fig.~\ref{fig:synthetic_images}) and rotate the camera from 170$^{\circ}$ to 190$^{\circ}$ in 5$^{\circ}$ steps. The positions were chosen to simulate the system being used by pedestrians.}{Synthetic images are rendered from $4\times4$ points and change the rotation by $5$ degrees as shown in \figref{fig:camera_position} because we aim that our system is used by pedestrians.}
\replaced[id=AP]{Overall, we generate 80 images per lighting condition, \ie 4,480 synthetic images in total. We show examples of the rendered images in \figref{fig:synthetic_images}.}
{We generate 80 synthetic images under one lighting condition, \ie 4,480 synthetic images in total.
Some rendered images are shown in \figref{fig:synthetic_images}.}

\subsection{Training Random Forest}\label{sec:train_rf}
\replaced[id=SS]{For feature matching, we use random forests because it is used for feature matching and training simplicity.}
{We use a random forest for feature matching because it can match them fast and the required dataset size for training is less than DNNs.} 
The random forest classifies a synthetic feature transformed by \NetName\ into its corresponding to synthetic feature cluster. Since synthetic feature clusters are related with a scene coordinate, correspondences between a image coordinate and a scene coordinate can be obtained.

However, random forest can mismatch a synthetic feature which is corresponding to non-synthetic feature clusters, \ie a noise feature, with one of the feature clusters.
To remove mismatches and reduce the time which RANSAC takes, we apply filtering before executing RANSAC.
A random forest estimates probabilities which input values are classified to a class. We ordered all matches by maximum probabilities, \ie the probabilities the feature is matched with the predicted synthetic feature clusters, and select the top 100 matches as an input to RANSAC.

\section{Evaluation}
\begin{table*}[t]
    \centering
	\caption{Result of evaluation for \NetName. From left column, matching accuracy (MA), position error (PE), orientation error (OE), and localization time are indicated.}
    \label{tb:result_transform}
    \tabcolsep=1em
    \begin{tabularx}{.85\linewidth}{ccc|NNNN}
    	\hline
        & & & \mathrm{MA} & \mathrm{PE} & \mathrm{OE} & \mathrm{Time}\\
        \hline
        
        \multirow{6}{*}{SIFT}        
        & \multirow{2}{*}{na\"{i}ve method} & mean   & 47.77\% & 19.75\ \mathrm{cm} & 26.01^\circ & 373.12\ \mathrm{ms} \\
        &                               & median & 57.58\% & 3.47\ \mathrm{cm} & 2.84^\circ & 375.79\ \mathrm{ms} \\
        & \NetName      & mean   & 58.81\% & 30.09\ \mathrm{cm} & 34.45^\circ & 344.51\ \mathrm{ms}\\
        & w/o whitening & median & 74.54\% & 5.75\ \mathrm{cm} & 4.39^\circ & 343.00\ \mathrm{ms}\\
        & \NetName     & mean   & \bm{83.44}\% & \bm{5.55}\ \mathrm{cm} & \bm{6.38}^\circ & \bm{340.50}\ \mathrm{ms} \\
        & w/ whitening & median & \bm{85.88}\% & \bm{1.89}\ \mathrm{cm} & \bm{1.64}^\circ & \bm{339.90}\ \mathrm{ms}\\
        \hline

		\multirow{6}{*}{SURF}
        & \multirow{2}{*}{na\"{i}ve method} & mean   & 62.33\% & 13.78\ \mathrm{cm} & 15.14^\circ & 321.39\ \mathrm{ms} \\
        &                               & median & 69.93\% & 3.35\ \mathrm{cm} & 2.65^\circ & 315.33\ \mathrm{ms} \\
        & \NetName      & mean   & 77.88\% & 6.32\ \mathrm{cm} & 6.83^\circ & 298.55\ \mathrm{ms} \\
        & w/o whitening & median & 80.52\% & 2.35\ \mathrm{cm} & 1.84^\circ & 300.80\ \mathrm{ms} \\
        & \NetName     & mean   & \bm{84.64}\% & \bm{5.07}\ \mathrm{cm} & \bm{5.45}^\circ & \bm{297.33}\ \mathrm{ms} \\
        & w/ whitening & median & \bm{86.95}\% & \bm{1.66}\ \mathrm{cm} & \bm{1.41}^\circ & \bm{297.79}\ \mathrm{ms} \\
        \hline

		\multirow{6}{*}{ORB}
        & \multirow{2}{*}{na\"{i}ve method} & mean   & 28.77\% & 7.28\mathrm{E}{+}9\ \mathrm{cm} & 62.82^\circ & 285.01\ \mathrm{ms} \\
        &                               & median & 21.11\% & 47.76\ \mathrm{cm} & 54.60^\circ & 284.38\ \mathrm{ms} \\
        & \NetName      & mean   & 36.60\% & \bm{2.29\mathrm{E}{+}5}\ \mathrm{cm} & 40.32^\circ & \bm{267.57}\ \mathrm{ms} \\
        & w/o whitening & median & 33.33\% & 5.97\ \mathrm{cm} & 5.23^\circ & \bm{267.26}\ \mathrm{ms} \\
        & \NetName     & mean   & \bm{44.68}\% & 1.47\mathrm{E}{+}11\ \mathrm{cm} & \bm{32.92}^\circ & 269.33\ \mathrm{ms} \\
        & w/ whitening & median & \bm{55.40}\% & \bm{5.48}\ \mathrm{cm} & \bm{4.20}^\circ & 269.72\ \mathrm{ms} \\
        \hline
    \end{tabularx}
\end{table*}

\replaced[id=AP]{We implemented \NetName\ on an Intel(R)  Core(TM) i7-6700K 4.00 GHz Desktop PC with 32 GB DDR4 RAM and an NVIDIA Geforce GTX 1080. We first evaluate if \NetName\ can improve the localization results on a database composed of synthetic features. Also, we will compare the performance of our approach with conventional CNN-based localization methods, such as PoseNet~\cite{Kendall_ICCV2015}, to show how our approach used to remove gap between synthetic and real images can improve localization accuracy.}
{In this section, we evaluate our system in order to \NetName\ can reduce the gap in \secref{sec:exp_transform}, and compare our system with a traditional camera localization method PoseNet \cite{Kendall_ICCV2015} in \secref{sec:exp_posenet}. Our computer spec is shown in \tbref{tb:spec}}


\subsection{\replaced[id=AP]{Effects of \NetName\ on Localization Accuracy}{Comparison between with and without Feature Transform}}
\label{sec:exp_transform}

\begin{figure}[t]
    \setlength{\imgwidth}{0.3\linewidth}
	\centering
    \begin{tabular}{ccc}
    	\includegraphics[width=\imgwidth]{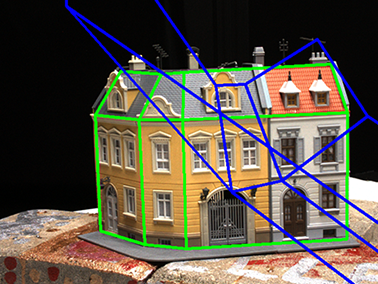}
        & \includegraphics[width=\imgwidth]{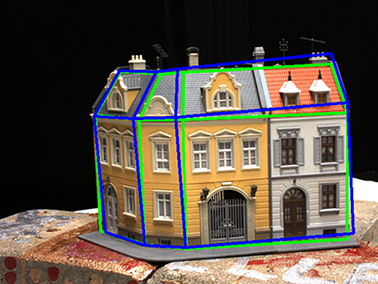}
        & \includegraphics[width=\imgwidth]{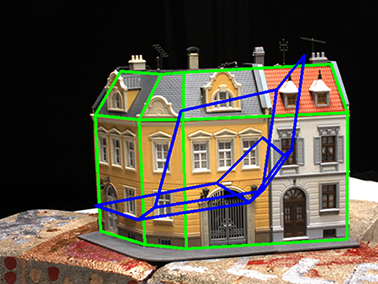}\\

		\includegraphics[width=\imgwidth]{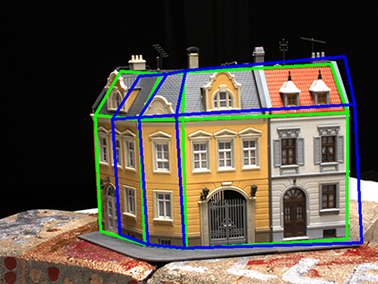}
        & \includegraphics[width=\imgwidth]{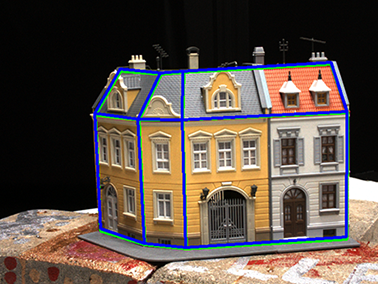}
        & \includegraphics[width=\imgwidth]{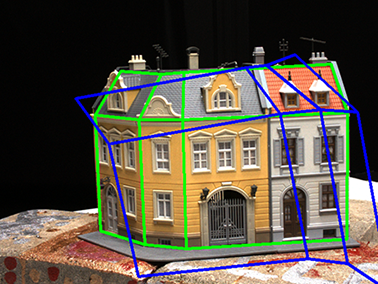}\\

		\includegraphics[width=\imgwidth]{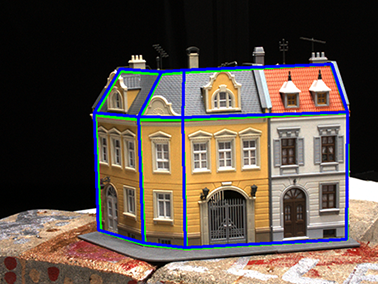}
        & \includegraphics[width=\imgwidth]{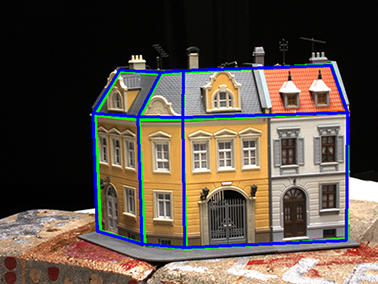}
        & \includegraphics[width=\imgwidth]{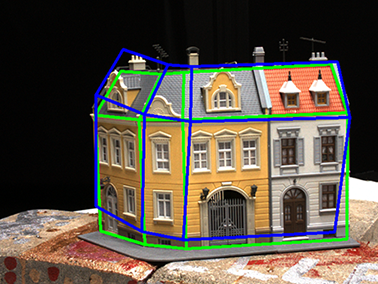}\\
    	SIFT & SURF & ORB\\
    \end{tabular}
    \caption{Localization result of na\"{i}ve method (top row), \NetName\ without whitening (middle row) and \NetName\ with whitening (bottom row). The green outline shows the ground truth outline of the building. The blue outline is the projection of the model given the estimated camera pose.}
    \label{fig:frame_feature}
\end{figure}

\begin{figure*}[t]
	\setlength{\imgwidth}{0.18\linewidth}
    \centering
    \begin{tabular}{ccccc}
	    \includegraphics[width=\imgwidth]{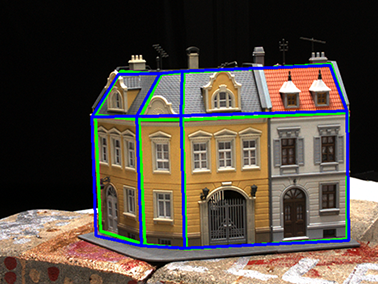}
	    & \includegraphics[width=\imgwidth]{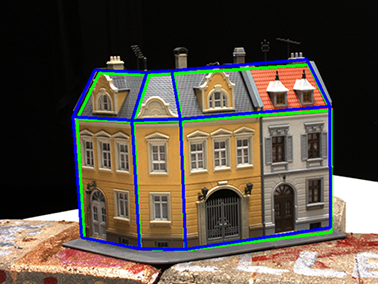}
	    & \includegraphics[width=\imgwidth]{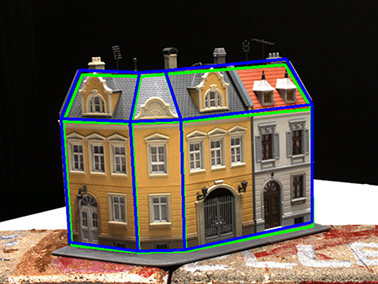}	
	    & \includegraphics[width=\imgwidth]{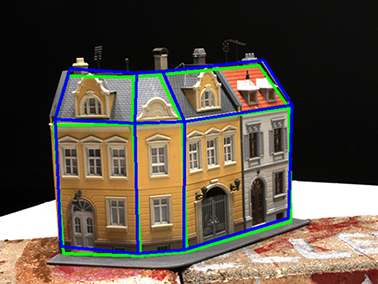}
    	& \includegraphics[width=\imgwidth]{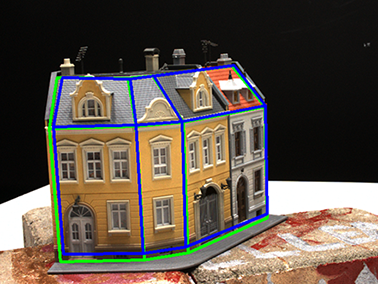}\\

		\includegraphics[width=\imgwidth]{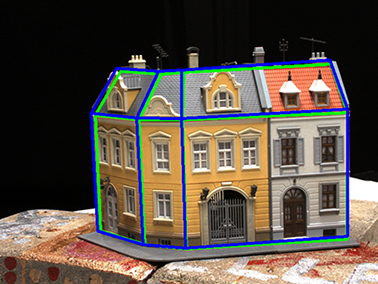}
    	& \includegraphics[width=\imgwidth]{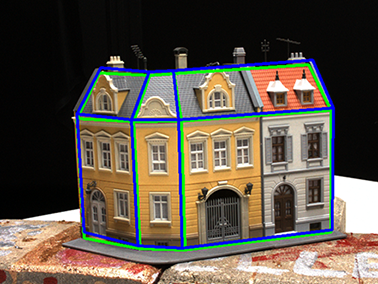}
   	 	& \includegraphics[width=\imgwidth]{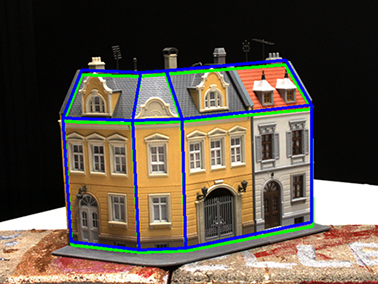}
    	& \includegraphics[width=\imgwidth]{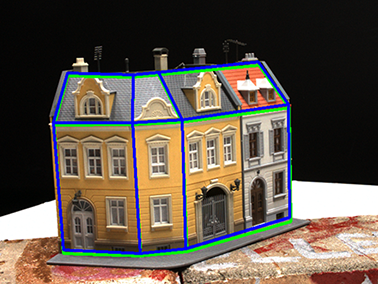}
    	& \includegraphics[width=\imgwidth]{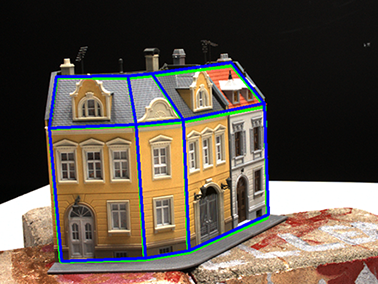}\\

		\includegraphics[width=\imgwidth]{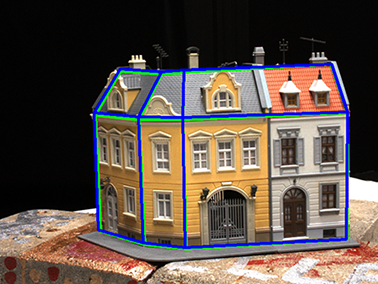}
    	& \includegraphics[width=\imgwidth]{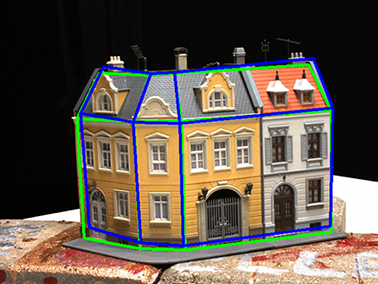}
    	& \includegraphics[width=\imgwidth]{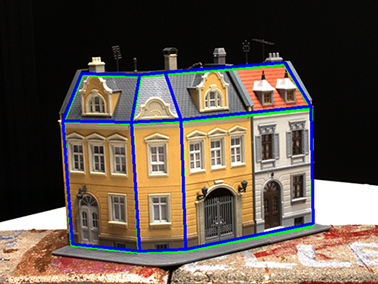}
    	& \includegraphics[width=\imgwidth]{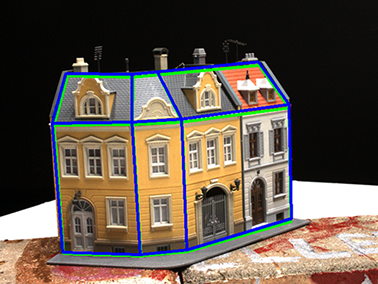}
    	& \includegraphics[width=\imgwidth]{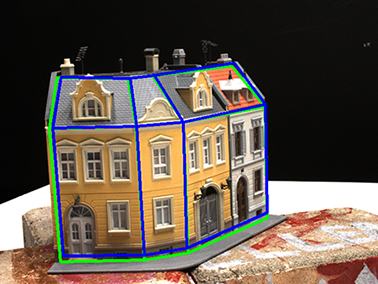}\\

		\includegraphics[width=\imgwidth]{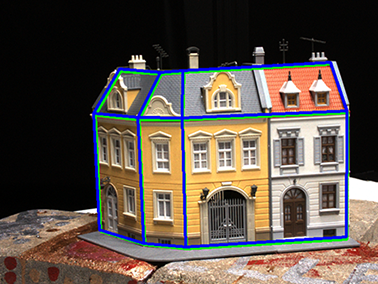}
    	& \includegraphics[width=\imgwidth]{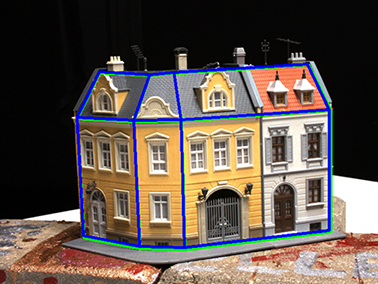}
    	& \includegraphics[width=\imgwidth]{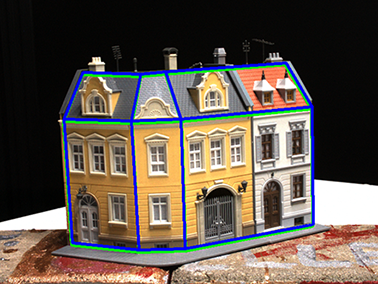}
    	& \includegraphics[width=\imgwidth]{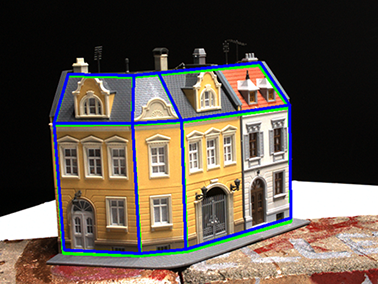}
   	 	& \includegraphics[width=\imgwidth]{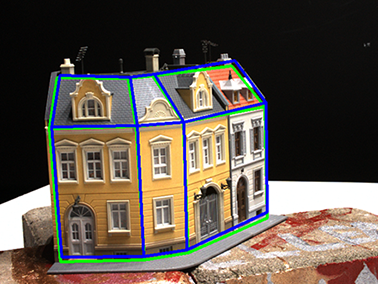}\\

		\includegraphics[width=\imgwidth]{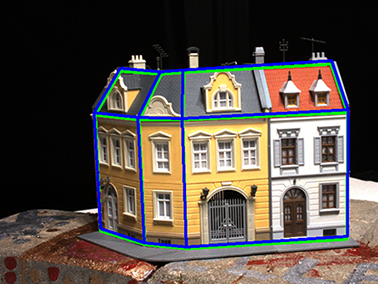}
    	& \includegraphics[width=\imgwidth]{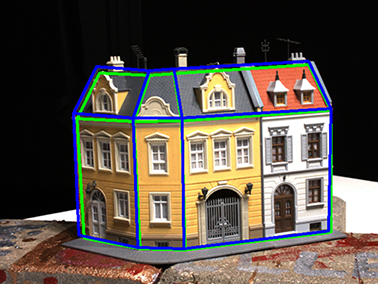}
   		& \includegraphics[width=\imgwidth]{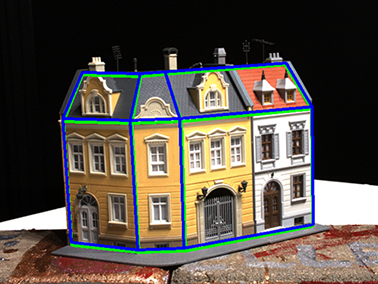}
    	& \includegraphics[width=\imgwidth]{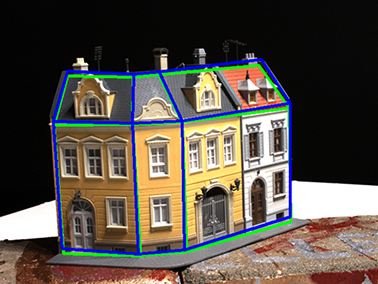}
    	& \includegraphics[width=\imgwidth]{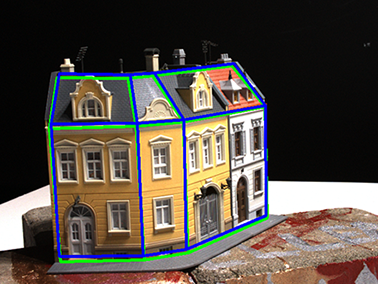}\\       
  	\end{tabular}
    \caption{Localization Result of \NetName\ with whitening using SIFT under various conditions. Columns represent camera viewpoints and rows represent lighting conditions.}
    \label{fig:frame_condition}
\end{figure*}

\added[id=AP]{To evaluate what effect \NetName\ as well as whitening has on the feature matching and localization accuracy, we compare \NetName\ with whitening, \NetName\ without whitening, and na\"{i}ve feature matching (directly matching a real feature with the synthetic feature database). We also evaluate how the chosen feature descriptor chosen affects the results. We compare \replaced[id=SS]{three}{3} commonly used descriptors: SIFT~\cite{Lowe_IJCV2004}, SURF~\cite{Bay_ECCV2006}, and ORB~\cite{Rublee_ICCV2011}. Since ORB is a binary feature descriptor, we split ORB features into float features by bits, \ie 256-dimensional features.}

\replaced[id=AP]{We evaluate all methods on 60 images from the DTU dataset~\cite{Jensen_CVPR2014}. The images were selected to match the view a pedestrian usually looks at the building. Each image has a resolution of 640 $\times$ 480 pixel and we split them into 5 groups for cross validation. We train \NetName\ with 4 groups and evaluate matching accuracy with the left out group.}{To confirm that \NetName\ can reduce the gap, we compare matching accuracy and the localization accuracy among \NetName\ with whitening, \NetName\ without whitening and \replaced[id=AP]{the na\"{i}ve approach}{naive method} (directly matching a real feature with \replaced[id=AP]{the synthetic feature database}{a synthetic feature}). We \replaced[id=AP]{selected}{prepare} 60 \deleted[id=AP]{real} images \replaced[id=AP]{from the DTU dataset. The images were selected to match the view a pedestrian would have looking at the building model.}whose size is $640 \times 480$ as test data and split them into 5 groups to do cross validation. These real images are taken from front of the scene object. These camera positions assumes pedestrians' view point.}
\deleted[id=AP]{For test image, we use DTU robot image dataset~\cite{Jensen_CVPR2014}. Although this has 392 images for a model, we use 60 images which are taken near floor plane.
This also assumes pedestrians' view point.
Since part of these real images are used also for training, we split them by five groups to execute 5-cross validation, \ie 48 images are used for training and the other 12 images for testing. }
\replaced[id=AP]{The na\"{i}ve method matches real features with the database using the same trained random forest as \NetName.}{For a naive method, we match raw real feature with synthetic feature clusters with the same random forest.}

\replaced[id=AP]{We judge a real feature as being correctly matched to its synthetic counterpart if the projection of the 3D location of the synthetic feature into the camera image, given the ground truth camera pose, is less than 3 pixel away from the detected real feature. This threshold is relatively small and is similar to ~\cite{Kurz_VISAPP2014}. We show the results of the matching accuracy in \tbref{tb:result_transform}.}
{To evaluate matching accuracy, we judge that matches whose reprojection error is less than 3 pixels are valid, otherwise invalid. The threshold 3 is approximately 0.5 \% of image width and approximately 0.6 \% of image height.}
\deleted[id=AP]{To evaluate localization accuracy, we compute the camera position error in centimeter and the camera orientation error in degree. According to \cite{Jensen_CVPR2014}, area range is $1\ \mathrm{m} \times 1\ \mathrm{m}$.}

\deleted[id=AP]{\tbref{tb:result_transform} shows the result.}
\replaced[id=AP]{\NetName\ improves the number of correctly matched features (MA) compared to the na\"{i}ve approach f}{F}or all feature descriptors\replaced[id=AP]{.}{, matching accuracy (MA) of \NetName\ is improved from that of the naive method}
This \deleted[id=AP]{result }indicates that \replaced[id=AP]{\NetName\ successfully converts real features into synthetic features.}{the real-to-synthetic feature transform enable to match a real feature with a synthetic feature.} \added[id=AP]{Applying whitening further improves the matching results for all descriptors. ORB features performed worst in all tests, while SURF features were more robust to the gap and performed best during na\"{i}ve matching. While this advantage persists for \NetName\ without whitening and with whitening, for \NetName\ with whitening SIFT features perform almost as well. \added[id=SS]{The performance of \NetName\ without whitening SIFT features is worse than na\"{i}ve matching. As discussed in \secref{sec:correspondence}, directly corresponding a real feature to a synthetic feature causes transformation a real feature into a synthetic feature in another feature clusters. As a result, \NetName\ decrease feature matching accuracy.} We show the localization results using the different features in \figref{fig:frame_feature}.}
\deleted[id=AP]{Moreover both of camera position error (PE) and orientation error (OE) of \NetName\ with whitening are less than the naive method, which suggests that the feature transform contributes to the camera localization accuracy.
Although the localization time is much slower than PoseNet, approximately 750 ms is enough to localize camera pose because camera localization methods are used only at the first frame or when lost tracking.}

\added[id=AP]{We evaluate the localization accuracy of the different methods. With the increasing ratio of correctly matched features we also observe improved localization accuracy. We show localization results of the na\"{i}ve approach, \NetName\ without whitening, and \NetName\ with whitening for SIFT features in \figref{fig:frame_feature}. The green outline shows the ground truth outline of the building, while the blue outline is the projection of the model given the estimated camera pose. \NetName\ with whitening is the only approach to recover an accurate camera pose. \NetName\ without whitening includes some outliers into which results in lower accuracy, while the na\"{i}ve approach fails to correctly estimate the pose due to a low number of correct matches. The DTU dataset covers an area of approximately 1 m $\times$ 1 m. Overall, using SURF features leads to the best and most stable results. Interestingly, although the correct matching ratio for SIFT features increases for \NetName\ without whitening compared to the na\"{i}ve approach, the localization accuracy decreases. However, it increases drastically for \NetName\ with whitening. This effect does not appear for SURF and ORB.}

\added[id=SS]{As stated in \secref{sec:intro}, lighting variation cause feature matching failure and decrease localization accuracy \cite{Mashita_ICAT2016}. Whereas, \NetName\ with whitening for SIFT can estimate the correct camera poses from real images under different camera positions and lighting conditions shown in \figref{fig:frame_condition}. Each columns represent a camera viewpoint and each rows represent a lighting condition. Thus, our approach is robust to variation of camera viewpoints and lighting conditions.}

\deleted[id=AP]{Comparing between \NetName\ with and without whitening, \NetName\ with whitening outperforms \NetName\ without whitening in terms of matching accuracy, camera position error and camera orientation error for SIFT and SURF.
Focusing on the mean of camera position error and camera orientation error, \NetName\ without whitening has much bigger errors than \NetName\ with whitening.
It suggests that \NetName\ without whitening sometimes fails localization extremely.
As we stated in \secref{sec:training_data}, \NetName\ without whitening transforms a real feature near the boundaries of the corresponding synthetic feature cluster and it causes failure of feature matching.
Actually, the mean of matching accuracy of \NetName\ without whitening is approximately 15 \% less than median of matching accuracy.
Therefore, whitening contributes to improve the accuracy of feature matching.}

\deleted[id=AP]{Figure~\ref{fig:frame_method} shows object frame predicted by \NetName\ with whitening, \NetName\ without whitening and the naive method using SIFT.
The blue lines represent a predicted object frame and the green lines represent ground truth object frame.
\NetName\ with whitening predicts true camera pose, but \NetName\ without whitening and the naive method fails camera localization.
Although \NetName\ without whitening matches 81.81\% features successfully, RANSAC includes invalid matchings as inliers.
This causes the failure of camera localizations.
Since the naive method cannot matching real features with synthetic features correctly, RANSAC   does not converge.}

\subsection{Comparison with PoseNet}\label{sec:exp_posenet}
\begin{table}[t]
	\centering
    \caption{Result of comparison with PoseNet. From left column, position error (PE), orientation error (OE), and localization time are indicated. 
    }
    \label{tb:exp_posenet}
	\begin{tabular}{ccc|nnn}
    \hline 
	Net & Image &  & \mathrm{PE} & \mathrm{OE} & \mathrm{Time} \\
	\hline 
	\multirow{4}{*}{PoseNet} & \multirow{2}{*}{real} &
	mean & 37.21\ \mathrm{cm} & 25.33^\circ & \bm{14.05}\ \mathrm{ms}\\
	&& median  & 37.17\ \mathrm{cm} & 25.69^\circ & \bm{13.17}\ \mathrm{ms}\\
	& \multirow{2}{*}{syn.} &
	mean & 3.87\ \mathrm{cm} & 10.87^\circ & 13.30\ \mathrm{ms}\\    
	&& median & 3.64\ \mathrm{cm} & 11.00^\circ & 13.05\ \mathrm{ms}\\    
	\hline 
	\NetName & \multirow{2}{*}{real} &
	                      mean & \bm{5.55}\ \mathrm{cm} & \bm{6.38}^\circ & 340.50\ \mathrm{ms}\\
	SIFT w/ whitening & & median & \bm{1.89}\ \mathrm{cm} & \bm{1.64}^\circ & 339.90\ \mathrm{ms}\\
    \hline 
	\end{tabular}
\end{table}

\replaced[id=AP]{We also compare our method with a CNN-based localization method on synthetic images. We trained PoseNet~\cite{Kendall_ICCV2015} on the same dataset that was used to generate the database. To verify the training quality we evaluate the PoseNet localization results on synthetic images that were not part of the training. PoseNet achieves high positional accuracy, while the rotation has an error of up to 11$^{\circ}$. However, when a real image is used instead, the positional and rotational accuracy degrades significantly. Overall, PoseNet performs worse than our approach. We show the results of PoseNet estimation in Table~\ref{tb:exp_posenet}. This indicates that the gap between simulated and real images also has a significant effect on CNN-based methods and correction with GAN-based methods is necessary. However, as training a GAN requires a large number of images it is not viable to simulate variable lighting conditions.}
{In this section, we compare the localization accuracy of our system with that of PoseNet.
For training of PoseNet, we use the synthetic images which are the same images used for the construction of our database.
Note that we train PoseNet without any real images.
Then, we test PoseNet using the real images which are the same images used for testing of our system.
Additionally, to clear the influence of the gap, we test PoseNet with synthetic images which are different from the synthetic images for training.}

\added[id=SS]{However, localization time of our method is slow than PoseNet. This is because feature description and feature matching with a random forest are time-consuming. Since camera localization is only required at the first frame of tracking and when tracking was lost. Thus, Our localization with the update rate of 350 ms is enough to camera localization. note that, our system run on CPU, which still have room for further optimization of localization time.}

\deleted[id=AP]{\tbref{tb:exp_posenet} shows the result.
The errors of PoseNet tested with real images are much bigger than those of PoseNet tested with synthetic images.
This shows that synthetic images can cause decrease of the localization accuracy.
PoseNet uses GoogLeNet architecture and adds three full connection layers to the last of GoogLeNet.
Since PoseNet is trained with synthetic images, GoogLeNet learn how to extract a feature form a synthetic image but not from a real image.
Thus, when inputting a real image, GoogLeNet cannot extract its feature.
We consider that this is why PoseNet trained synthetic images fails to localize camera pose.}

\section{Conclusion and Future Work}
\replaced[id=AP]{In this paper we presented a new approach for robust camera localization in varying lighting conditions. We use a feature database generated from synthetic images that simulate the appearance of the scene under different lighting conditions. Sicne the synthetic data does not perfectly match the scene appearance, it is necessary to overcome the appearance gap between simulation and reality. 

We introduce \NetName, an autoencoder-like network that converts feature descriptors extracted from real images into descriptors which similar to the descriptors that would be extracted from synthetic images generated under the same conditions. We also use whitening process to improve the matching ratio between real and synthetic features. Our pipeline successfully improves the matching ratio between real images and the feature database. Our experimental results show that the appearance gap clearly affects CNN-based localization methods. Therefore, methods trained only on synthetic images fail to correctly localize the camera.}
{We have illustrated lighting robust camera localization system.
To be robust to lighting conditions, generating synthetic images is one of the solution.
However, illumination simulation cannot perfectly render synthetic images likely to real images.
The gap between real images and synthetic images can cause decrease of localization accuracy.
To solve the gap problem, we have proposed \NetName\ which transforms real feature into synthetic feature.
For preparation of the training data of \NetName, we utilize whitening to associate a real feature and a synthetic features in an overlapped feature space.}


\added[id=SS]{Since \NetName\ converts a real feature into a synthetic feature which might not always matched with feature database. For the future work, we plan to improve our system so that it can convert a real feature to a feature that is easily matched to feature database. This improves not only the accuracy of camera localization but also localization time because simpler and faster feature matching can be achieved without random forests.}

\deleted[id=AP]{Our results indicate that gap between real and synthetic images causes decrease of localization accuracy with camera localization system with end-to-end CNN, and \NetName\ can reduce the influence of the gap.
To further improve the accuracy, we will aim to transform a real feature to feature which is estimated to be matched easily, whereas \NetName\ transforms real feature into nearest synthetic feature in an overlapped feature space.
Since this transformation contributes the localization accuracy directly, further improvements are expected.}

\section*{Acknowledgement}
Part of this work was supported by JSPS KAKENHI Grant Number JP16H02858 and JP16K16100.




\end{document}